%% file: main.tex
\documentclass[10pt,twocolumn,letterpaper]{article}

\usepackage{cvpr}              

\usepackage{multirow}
\usepackage{booktabs}
\usepackage{bbding}
\usepackage{url}

\input{preamble}

\definecolor{cvprblue}{rgb}{0.21,0.49,0.74}
\usepackage[pagebackref,breaklinks,colorlinks,citecolor=cvprblue]{hyperref}

\def\paperID{15445} 
\def\confName{CVPR}
\def\confYear{2024}

\title{Rethinking Image Editing Detection in the Era of Generative AI
Revolution}

\author{Zhihao Sun$^{1,2}$, Haipeng Fang$^{1,2}$, Xinying Zhao$^{1,2}$, Danding Wang$^{1,2}$, Juan Cao$^{1,2}$\\
$^{1}$Institute of Computing Technology, Chinese Academy of Sciences\\
$^{2}$University of Chinese Academy of Sciences\\
}

\begin{document}

\twocolumn[{
    \renewcommand\twocolumn[1][]{#1}
    \maketitle
    \begin{center}
    \centering
    \includegraphics[width=\linewidth]{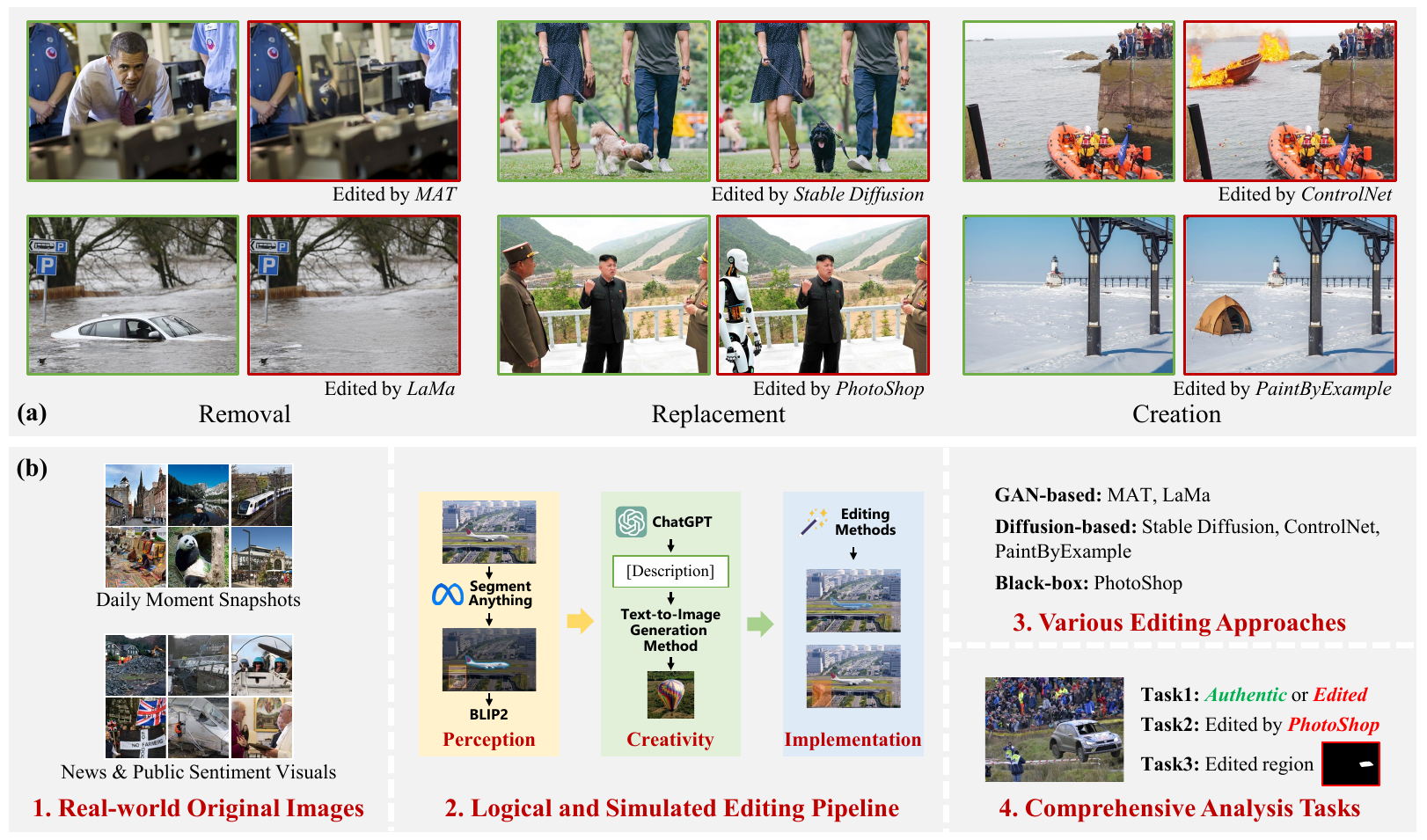}
    \captionof{figure}{GRE is a large-scale generative regional editing dataset with several distinct advantage, briefly highlighted in (b). We showcase some edited image with different editing approaches and various editing types in (a), along with their corresponding original images.}
    \label{fig:head}
    \end{center}
}]

\input{sec/0_abstract}    
\input{sec/1_intro}
\input{sec/2_related}
\input{sec/3_construct}

\input{sec/4_analysis}
\input{sec/5_conclusion}

{
    \small
    \bibliographystyle{ieeenat_fullname}
    \bibliography{main}
}


\end{document}

%% file: preamble.tex
\usepackage[dvipsnames]{xcolor}


%% file: sec/0_abstract.tex
\begin{abstract}

The accelerated advancement of generative AI significantly enhance the viability and effectiveness of generative regional editing methods. This evolution render the image manipulation more accessible, thereby intensifying the risk of altering the conveyed information within original images and even propagating misinformation. Consequently, there exists a critical demand for robust capable of detecting the edited images. However, the lack of comprehensive dataset containing images edited with abundant and advanced generative regional editing methods poses a substantial obstacle to the advancement of corresponding detection methods.

We endeavor to fill the vacancy by constructing the GRE dataset, a large-scale generative regional editing dataset with the following advantages: 1) Collection of real-world original images, focusing on two frequently edited scenarios. 2) Integration of a logical and simulated editing pipeline, leveraging multiple large models in various modalities. 3) Inclusion of various editing approaches with distinct architectures. 4) Provision of comprehensive analysis tasks. We perform comprehensive experiments with proposed three tasks: edited image classification, edited method attribution and edited region localization, providing analysis of distinct editing methods and evaluation of detection methods in related fields. We expect that the GRE dataset can promote further research and exploration in the field of generative region editing detection.

\end{abstract}

%% file: sec/1_intro.tex
\section{Introduction}
\label{sec:intro}

Diffusion models spark an AI generation revolution in the field of computer vision, demonstrating remarkable performance in various task scenarios such as visual content generation and controllable editing \cite{rombach2022high, saharia2022photorealistic, yu2022scaling, zhang2021ernie}. However, the advancement of generative technologies also facilitate the dissemination of malicious fake information. While detection methods for image generation\footnote{In this paper, the term \textbf{image generation} specifically refers to instances where all pixels in an image are generated. Conversely, the term \textbf{regional editing} denotes the modification of only a portion of the pixels based on the original image, and this is also referred to as image manipulation in some related studies.} have gathered widespread attention and research, there remains a gap in the study of generative regional editing detection. 

In contrast to the challenging precise control in the full image generation techniques, local editing methods exhibit greater flexibility, which enable the modification of specific content in the original image \cite{rombach2022high, zhang2023adding, yang2023paint}, potentially altering the conveyed information. Moreover, compared to traditional manual manipulation using tools like PhotoShop \cite{photoshop}, the process of generative regional editing is more convenient and user-friendly for non-professionals, while still achieving high-quality editing results. Figure \ref{fig:head} (a) showcases the performance of several representative generative regional editing methods, illustrating the difficulty in distinguishing between authentic and edited images. In the present day, we can indeed assert that ``Seeing is not always believing." \cite{lu2023seeing} Therefore, the detection capabilities of generative regional editing merit our attention.

In this paper, we construct a novel large-scale dataset named \textbf{GRE} (Generative Regional Editing) with three tasks for generative regional editing analysis. We carefully evaluate the existing detection methods across related different domains using the GRE dataset. Extensive experiments and in-depth analysis demonstrate that this larger and more comprehensive dataset significantly enhances the development of detection methods for generative editing. Specifically, GRE dataset offers several distinct advantages over existing related datasets, which are listed below.

\noindent \textbf{(1) Real-world Original Images.} What types of images are most susceptible or frequently tampered in real world scenarios? We summarize them into two typical scenes: Daily Moment Snapshots capturing individual perspectives, and News \& Public Sentiment Visuals reflecting public perspectives. For these two pivotal classes, we gather the original images for our GRE dataset. These images exhibit diversity across multiple dimensions, including objects, content and scenes, laying the foundation for the subsequent operations of abundant editing intents. Simultaneously, we place emphasis on the diversity of image resolutions within the dataset, with resolutions ranging from 480P to 2K.

\noindent \textbf{(2) Logical and Simulated Editing Pipeline.} Previously, small-scale regional editing datasets ensured logical coherence through manual manipulation, while larger datasets struggled to maintain logical consistency through naive automated editing pipeline. To ensure logical coherence in editing (\textit{e.g.}, preventing the appearance of a dog in the sky), semantic richness in editing, data scale and scalability, we integrate multiple awesome large models in various modalities to construct a complete image editing pipeline including perception, creativity and implementation. In addition to obtaining edited images and corresponding region annotations, we document the intent behind the editing process in textual form, which serves to detection and analysis of editing intent, and enhancing interpretability and understanding of the regional editing detection.

\noindent \textbf{(3) Various Editing Approaches.} Since the remarkable performance of diffusion models \cite{DDPM, Score-based} in image generation tasks, numerous editing methods based on the principles of diffusion have emerged. While these methods share the same underlying principles, there are variations in network architecture design, editing control mechanisms, and other aspects. Whether these differences pose challenges to the generalization of region editing detection algorithms is a question worth exploring. Additionally, despite most GAN-based methods exhibit limitations in their generative capabilities, often restricted to object removal, these methods still hold value in specific scenarios. Moreover, due to the differences in principles, achieving generalization in detection for both GAN-based and diffusion-based editing methods is challenging. We select a variety of representative editing methods, including GAN-based, diffusion-based and black-box approaches, as well as different editing control mechanisms, for thorough investigation. 

\noindent \textbf{(4) Comprehensive Analysis Tasks.} According to the real application scenario, we provide annotations at multiple levels and propose three tasks: 1) Edited Image Classification, distinguishes whether an image is edited or not. 3) Edited Method Attribution, identifies the editing method used in an edited image. 2) Edited Region Localization, localizes manipulated areas within edited images. The pixel-level localization task, although more challenging, is meaningful in finding edited elements within a visually rich edited image. 

%% file: sec/2_related.tex
\section{Related Work}
\label{sec:related}

\begin{table*}[t]
    \centering
    \resizebox{\linewidth}{!}{
        \setlength{\tabcolsep}{2.5mm}{
            \begin{tabular}{lcccccccl}
                \toprule[1.3pt]
                \multirow{2}[1]{*}{\textbf{Dataset}} & \multicolumn{2}{c}{\textbf{Dataset Scale}} & \multicolumn{2}{c}{\textbf{Original Image}} & \multicolumn{3}{c}{\textbf{Generative Editing Approaches}} & \multirow{2}[1]{*}{\textbf{Pipeline}} \\
                \cmidrule(r){2-3} \cmidrule(r){4-5} \cmidrule(r){6-8} 
                
                 & Edited Images & Generative Ratio(\%) & Daily & News & GAN-based & Diffusion-based & Black-box & \\
                \midrule
                Columbia\cite{shi2000normalized} & 180 & 0.0 & \Checkmark & \XSolidBrush & \XSolidBrush & \XSolidBrush & \XSolidBrush & Random\\
                CASIAv1\cite{dong2013casia} & 920 & 0.0 & \Checkmark & \XSolidBrush & \XSolidBrush & \XSolidBrush & \XSolidBrush & Manual \\
                CASIAv2\cite{dong2013casia} & 5,063 & 0.0 & \Checkmark & \XSolidBrush & \XSolidBrush & \XSolidBrush & \XSolidBrush & Manual \\
                Coverage\cite{wen2016coverage} & 100 & 0.0 & \Checkmark & \XSolidBrush & \XSolidBrush & \XSolidBrush & \XSolidBrush & Manual \\
                NIST16\cite{guan2019mfc} & 564 & 36.9 & \Checkmark & \XSolidBrush & \Checkmark & \XSolidBrush & \XSolidBrush & Manual \\
                DEFACTO\cite{mahfoudi2019defacto} & 149,587 & 16.7 & \Checkmark & \XSolidBrush & \Checkmark & \XSolidBrush & \XSolidBrush & Random \\
                IMD20\cite{novozamsky2020imd2020} & 2,010 & 0.0 & \Checkmark & \XSolidBrush & \XSolidBrush & \XSolidBrush & \XSolidBrush & Manual \\
                \textbf{GRE (Ours)} & \textbf{228,650} & \textbf{100.0} & \Checkmark & \Checkmark & \Checkmark & \Checkmark & \Checkmark & \textbf{Simulated} \\

                \bottomrule[1.3pt]
            \end{tabular}
        }
    }
    \caption{Summary of various regional editing detection datasets. GRE surpasses any other dataset both in scale and diversity. The carefully designed simulated editing pipeline is also a significant advantage.}
    \label{table:datasets}
\end{table*}

\subsection{Generation and Manipulation Datasets}

\noindent \textbf{Image Generation.} Recently, there has been a growing emphasis on the detection of generative images, leading to the introduction of numerous benchmarks such as DeepArt \cite{wang2023benchmarking}, IEEE VIP Cup \cite{verdoliva2022}, DE-FAKE \cite{xu2023exposing}, and CiFAKE \cite{bird2023cifake}, along with the million-scale benchmark provided by GenImage \cite{zhu2023genimage}. However, the generative images within these datasets are primarily suitable for image-level generation detection tasks. They do not fully meet the requirements for the edited region localization task. Creating datasets specifically for the generative regional editing detection task incurs higher costs, and its pixel-level automated editing process is more complex compared to image-level generation.

\noindent \textbf{Regional Image Editing.} Detecting tampering or editing regions in an image is a longstanding challenge. Table \ref{table:datasets} provides a summary of the scale, image source, and editing approaches of existing datasets, including Columbia \cite{shi2000normalized}, CASIA \cite{dong2013casia}, Coverage \cite{wen2016coverage}, NIST16 \cite{guan2019mfc}, DEFACTO \cite{mahfoudi2019defacto} and IMD20 \cite{novozamsky2020imd2020}, which are widely used and recognized. Among these datasets, only the DEFACTO dataset includes a relatively extensive collection of generative edited image data. Other datasets predominantly include early non-generative forms of editing (e.g., simple splice and copy-move). However, the generative editing methods employed in DEFACTO dataset are limited, and the automated editing pipeline used is relatively simple. This editing pipeline leaves noticeable traces of automation, resulting in significant generalization issues for models trained on the dataset.

\subsection{Generative Regional Editing Methods}

\noindent \textbf{Diffusion-based methods.} The emergence of diffusion models has truly propelled generative editing methods to outperform operation sequences dominated by manual interventions, both in terms of convenience and effectiveness. Stable Diffusion \cite{rombach2022high} represents an advanced text-to-image diffusion model capable. The inclusion of simple mask replacement operations during the inference process enables targeted region editing. ControlNet \cite{zhang2023adding} introduces innovative modules that enable the control of pre-trained large-scale diffusion models to accommodate additional input conditions. PaintbyExample \cite{yang2023paint} explores exemplar-guided image editing rather than language-guided image editing, enabling even more precise control over the editing process.

\noindent \textbf{GAN-based methods.} However, we must also acknowledge the significant performance improvements in GAN-based image editing methods that have occurred in recent times. MAT \cite{li2022mat} customizes an inpainting-oriented transformer block, in which the attention module aggregates non-local information exclusively from partially valid tokens, as indicated by a dynamic mask. This approach demonstrates remarkable effectiveness in addressing extensive inpainting challenges. LaMa \cite{suvorov2022resolution} optimizes the intermediate feature maps of a network by minimizing a multi-scale consistency loss during inference. This approach adeptly handles the issue of lacking detail present at higher resolutions, resulting in improved visual quality.

%% file: sec/3_construct.tex
\begin{figure*}[th]
    \centering
    \includegraphics[width=\linewidth]{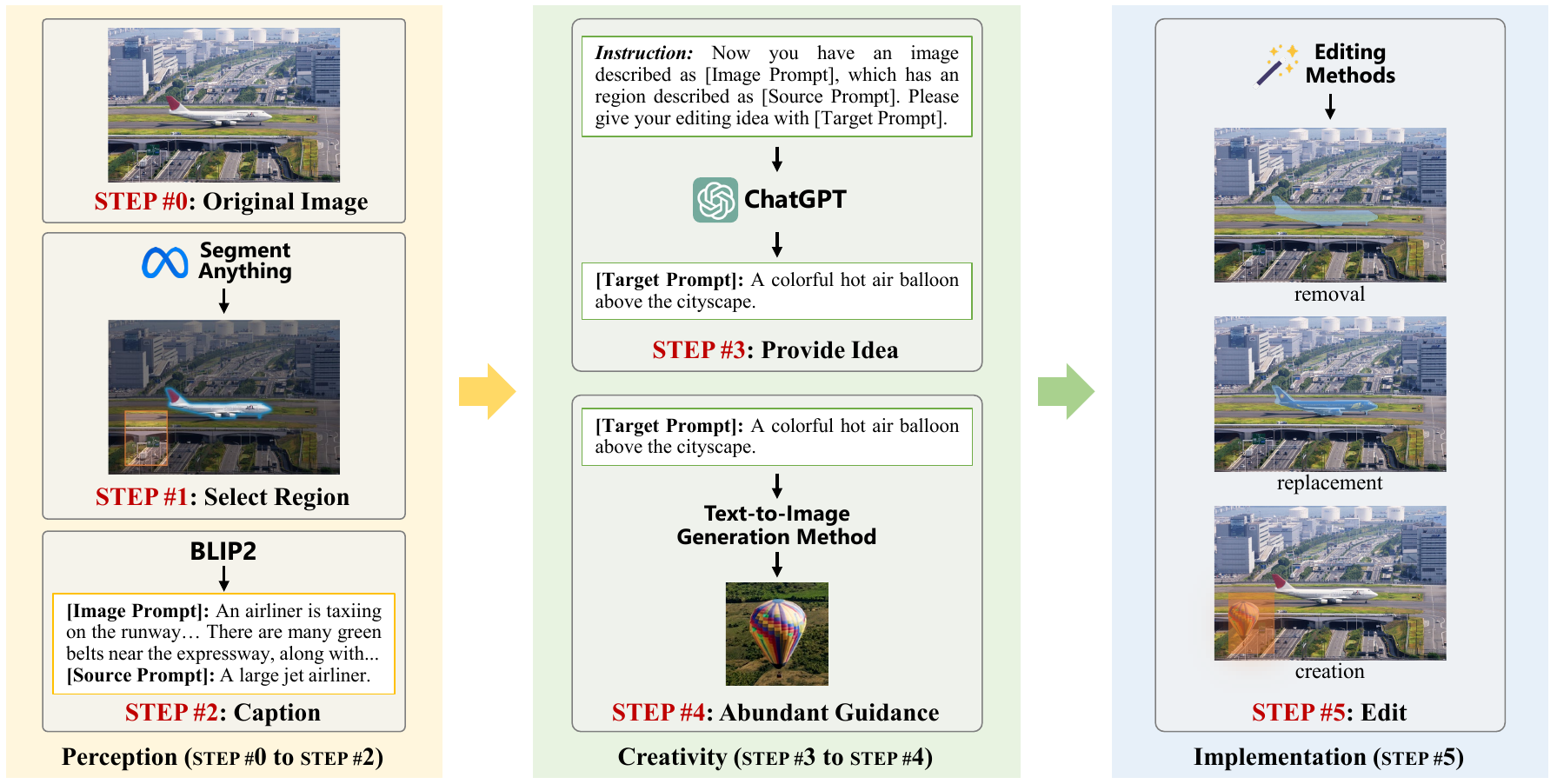}
    \caption{\textbf{Illustration of our logical and simulated pipeline for regional editing.} Simulating the real editing workflow, the process including three crucial components: \textbf{perception}, \textbf{creativity} and \textbf{implementation}. With the assistance of various large models in multiple modalities, this pipeline ensures the logical coherence and imperceptibility of editing operations.}
    \label{fig:pipeline}
\end{figure*}

\section{GRE Construction}

Most of existing image generation datasets only contain full image generated samples, without considering the common scenario of regional editing within images. Most previous regional editing datasets only contain manipulation without the participation of generative models, and the creation processes lack consideration of logical rationality and semantic diversity. On the contrast, our proposed GRE dataset provides various generative regional editing approaches, and defines three tasks (\textit{i.e.} edited image detection, edited region localization and editing method attribution) with a total of 228K images. We design a automated editing pipeline assisted by multiple large models with different modalities, capable of performing logically consistent editing operations and recording a diverse range of editing intents. We compare our GRE with other public regional editing datasets, as detailed in Table \ref{table:datasets}. Over all the comparison items listed in the table, our dataset outperforms others in both scale and diversity.

\subsection{Original Data Collection}

In the context of the internet, we categorize images frequently selected for tampering or editing into two primary classes: \textit{Daily Moment Snapshots} and \textit{News \& Public Sentiment Visuals}. Building upon this conceptual framework, we gather abundant original data to enhance diversity across dimensions such as scenes, content, and resolution.

\noindent \textbf{Daily Moment Snapshots} comprises user-shared pictures capturing daily life scenes and sharing moments, depicting the ordinary and personal aspects of individuals' lives. COCO \cite{lin2014microsoft} and Flickr2K \cite{timofte2017ntire} collected images from Flickr \cite{flickr}, comprising photographs uploaded by amateur photographers with searchable keywords, including 40 scene categories. Similarly, DIV2K \cite{agustsson2017ntire} and SR-RAW \cite{zhang2019zoom} gathered high-resolution images from a diverse set of websites and cameras, capturing snapshots of various moments and abundant contents. We select original data from these datasets, where the resolutions range from 480P to 2K.

\noindent \textbf{News \& Public Sentiment Visuals} includes visuals intricately linked to current events, news, or public sentiment, fostering broader discussions and sparking the attention of a larger audience. VisualNews \cite{liu2020visual} is a benchmark designed for the news image caption task, consisting of a large-scale collection of news images and associated metadata. The dataset was sourced from prominent news outlets such as BBC, USA Today, and The Washington Post, among others. From this dataset, we specifically select news illustrations with resolutions exceeding 720P and possessing rich content as the original images. 

\subsection{Regional Editing Pipeline}

To simulate the image editing process in real-world scenarios and ensure logical coherence in edited content, we design the editing pipelines assisted by multiple large models of different modalities, as illustrated in Figure \ref{fig:pipeline}. This pipeline primarily consists of three pivotal components. (1) Perception, which involves selecting the region to be edited and understanding the original image content. (2) Creativity, which involves determining the editing goal, and gathering corresponding textual descriptions and image examples (the guidance inputs for subsequent editing). (3) Implementation, which entails selecting the required guidance, employing various editing methods for multiple iterations of image editing and filtering the optimal result.

\begin{figure}[t]
    \centering
    \includegraphics[width=\linewidth]{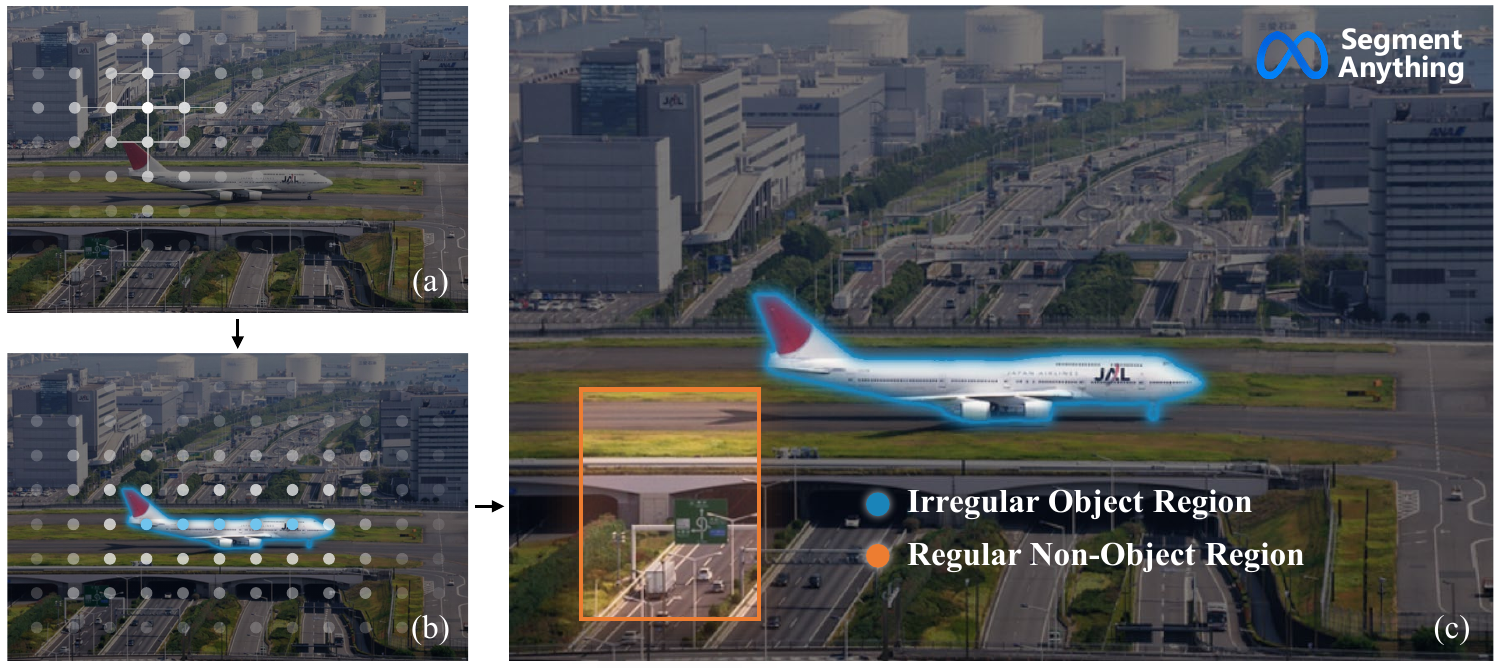}
    \caption{\textbf{Illustration of point-based SAM segmentation.} (a) Dense grid of points as the input for SAM. (b) Selection of an object if multiple points yield similar segmentation masks. (c) Outside the object region, selection of non-object regions.}
    \label{fig:sam}
\end{figure}

\subsubsection{Perception of Original Image}

\textbf{What is the content presented in the original image?} The first crucial component of the pipeline is to achieve perception of the original image. In this component, we aim to comprehend the image and select editing regions that are diverse and reasonable for subsequent editing. In real-world scenarios, edited regions can be broadly categorized into two types: object regions and non-object regions. For the former, editing operations such as removal or replacement can be performed, while for the latter, operations involve creating content that is not present in the original image. 

To simulate the selection of objects, we employ an advanced semantic segmentation model SAM \cite{kirillov2023segment} to obtain precise object region masks, as illustrated in Step \#1. SAM can achieve point-based segmentation. Therefore, we utilize a dense grid of points, as illustrated in Figure \ref{fig:sam} (a), to guide SAM for multiple region predictions. For an object or region with clear semantic meaning, it should be selected by at least two points and produce similar masks. We use this criterion to filter regions with complete semantic meaning. Conversely, outside these regions, there is a high probability of being background areas with no clear semantic meaning. In these cases, we use randomly sized rectangular regions to select these areas. We employ constraints related to size and the number of connected components to eliminate fragmented and meaningless segments. Consequently, we obtain irregular object region masks and regular non-object region masks, denoted as \texttt{\small [Region Mask]}, which is the most crucial guidance for subsequent editing process. 

We employ the large-scale visual-text model BLIP2 \cite{li2023blip} for the recognition of specified content in Step \#2. We aim for BLIP2 to provide a detailed description of the original image, referred to as \texttt{\small [Image Prompt]}. Subsequently, we crop the selected region with bounding boxes enlarged by 1.3x. We expect BLIP2 to provide a description of the original object or content within that region, denoted as \texttt{\small [Source Prompt]}. Finally, we analyze the coarse-grained position of the selected region in the image (using combinations such as center, top, bottom, left and right) and incorporate this information with the \texttt{\small [Source Prompt]}.

\subsubsection{Creativity of Editing}

\textbf{What is the idea and target of the editing?} In the real world, common editing types can be summarized as removal, replacement, and creation. Among these, removal is the most straightforward to establish, requiring only the \texttt{\small [Region Mask]} obtained in the earlier steps. However, for achieving the other editing types, the preparation of corresponding guidance that can describe the editing idea and purpose becomes essential.

ChatGPT \cite{chatgpt}, developed by OpenAI upon InstructGPT \cite{ouyang2022training}, is an excellent advisor to generate innovative editing ideas. We utilize carefully designed instruction format\footnote{\textbf{Instruction Format:} Now you have an image described as image prompt: \texttt{\footnotesize [Image Prompt]}, which has an region described as source prompt: \texttt{\footnotesize [Source Prompt]}. This area may be an object or an empty background area. Now you need to edit or tamper with this area. You can make creative edits out of fun, or you can change the original meaning of this image out of any intent. Please give your editing ideas and describe the expected target of this region as target prompt: [Target Prompt].} to inform ChatGPT about the content of the original image and the content of the selected region for editing. We hope that it can provide diverse and realistic editing ideas that align with real-world logic in Step \#3. The required text description of the editing target, \texttt{\small [Target Prompt]}, can be extracted from its response. We leverage the currently best open-source text-to-image generation model, Stable Diffusion XL \cite{podell2023sdxl}, to translate the text description into image examples \texttt{\small [Target Example]} in Step \#4. This serves as a different form of guidance needed for the subsequent editing process. It's essential to clarify that the target examples generated in this step do not belong to the final dataset, they are merely the guidance generated by the intermediate steps.

The potential improvement of editing detection performance through the incorporation of editing intent is worthwhile investigation. Considering that existing datasets do not annotate the editing intent for each edited image, we aim for ChatGPT to leverage the previously obtained \texttt{\small [Image Prompt]}, \texttt{\small [Source Prompt]}, and \texttt{\small [Target Prompt]} to analyze the intent behind each editing task. Subsequently, this \texttt{\small [Editing Intent]} is utilized as an annotation for each edited image.

\begin{table}[t]
\centering
\resizebox{\linewidth}{!}{
\begin{tabular}{llll}
    \toprule
    \textbf{Method} & \textbf{Architecture} & \textbf{Guidance} \\
    \midrule
    MAT & GAN & \texttt{\footnotesize [Region Mask]} \\
    LaMa & GAN & \texttt{\footnotesize [Region Mask]} \\
    StableDiff & Diffusion & \texttt{\footnotesize [Region Mask]},\texttt{\footnotesize [Target Pro.]} \\
    ControlNet & Diffusion & \texttt{\footnotesize [Region Mask]},\texttt{\footnotesize [Target Pro.]} \\
    PaintEx & Diffusion & \texttt{\footnotesize [Region Mask]},\texttt{\footnotesize [Target Pro.],\texttt{\footnotesize [Target Ex.]}} \\
    PhotoShop & Black-box & \texttt{\footnotesize [Region Mask]},\texttt{\footnotesize [Target Pro.]} \\
    
    \bottomrule
\end{tabular}
}
\caption{The distinctive characteristics of representative generative regional editing methods.}
\label{table:gen_methods}
\end{table}

\subsection{Various Generative Regional Editing Methods}

We have gathered comprehensive guidance information for region editing, including a precise binary mask indicating the editing region \texttt{\small [Region Mask]}, textual descriptions indicating the editing target \texttt{\small [Target Prompt]}, and image examples providing visual references for the editing target \texttt{\small [Target Example]}. These pieces of information offer diverse guidance for generative region editing methods, enabling end-to-end region editing.

Some works in image generation detection and attribution proposed and analyzed various generative methods from different perspectives, highlighting that different methods leave distinct traces and fingerprints \cite{yang2023fingerprints}. Moreover, there is a noted poor generalization of detection models across data generated by different methods. To ensure diversity in edited images within our GRE dataset and to provide a reasonable benchmark for generalization evaluation, we have chosen six editing methods to complete the final component in the pipeline, implementation. These six editing methods include MAT, LaMa, Stable Diffusion V2.0 (SD-V2.0), ControlNet, PaintByExample (PaintEx), and PhotoShop, which has introduced Generative AI functionality. Their characteristics are summarized in Table \ref{table:gen_methods}. 

For each original image, we employ all white-box methods to generate corresponding edited images. However, due to the manual intervention required in the generative editing process within PhotoShop, we select only a subset of images for PhotoShop editing. When using the three diffusion models in the above-mentioned editing methods, we incorporate diverse inference steps, randomly selecting the number of steps from the set $[20, 30, 50, 100]$ for each inference. Considering the variable quality of images generated by the diffusion-based model, multiple images are generated for each case. Subsequently, we choose the image with higher textual faithfulness based on the CLIP score \cite{radford2021learning}. Finally, we simulate real-world scenarios by introducing perturbations to the edited images, involving random combinations of different compression algorithms and noise addition algorithms, among other post-processing operations.

\begin{figure}[t]
    \centering
    \includegraphics[width=\linewidth]{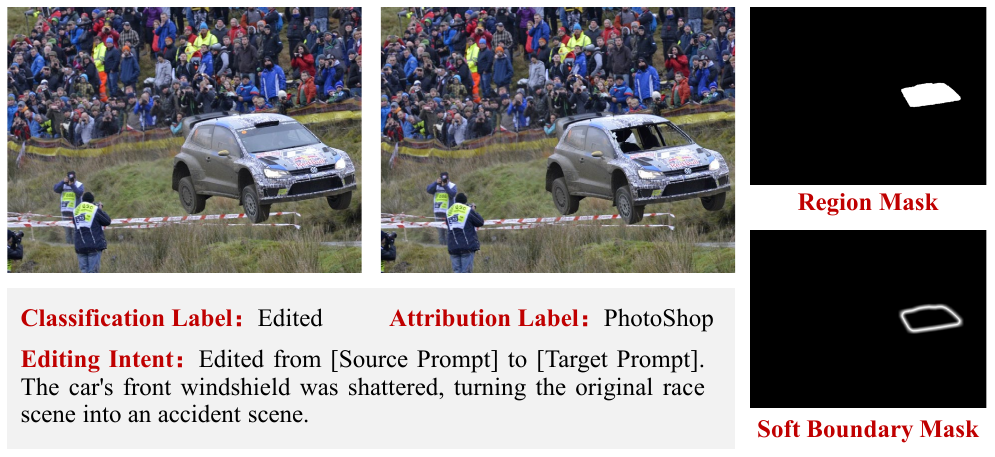}
    \caption{Examples showcasing the rich provided annotations.}
    \label{fig:annotations}
    \vspace{-1em}
\end{figure}

\subsection{Annotations}

For each image in the GRE dataset, we provide multiple annotations, including a $2$-way classification label (authentic or edited), an $n$-way classification label for the image (authentic or specific editing methods), and a binary segmentation mask (authentic region or edited region).

Furthermore, compared to the original images, all edited images have additional annotations shown as Figure \ref{fig:annotations}. After each editing operation, we obtain a textual description \texttt{\small [Intent Description]} of the editing intent summarized by ChatGPT. Several methods in the field of image tampering detection have validated that supervision on the boundaries of edited regions can enhance detection performance. Therefore, inspired by the boundary sliding mechanism in SAFL-Net \cite{sun2023safl}, we derived soft boundary masks.

%% file: sec/4_analysis.tex
\section{Generative Regional Editing Analysis}

\subsection{Evaluation on GRE Benchmark}

\subsubsection{Benchmark Settings}

\noindent \textbf{Dataset Partition.} For each original image collected in GRE, we employ all white-box methods to generate corresponding edited images, resulting in a dataset with a distribution from $1$ (authentic) to $n$ (edited). Consequently, we group images edited with the same method into a subset, while all original images formed the authentic subset. To ensure data uniformity and prevent data leakage, we initially partition the subset of authentic images into training, validation, and test sets in ratio of $8:1:1$. the division of each edited subset remains consistent with the authentic subset. In other words, if an original image is in the test set, all images edited from it also belong to the test set, ensuring exclusion from the training set.

\noindent \textbf{Task 1. Edited Image Classification.} This corresponds to a $2$-ways image-level classification task designed for distinguishing between authentic and edited images. We design the evaluation protocol to analyze the distinctions among various editing methods and assess the generalization performance of different detection methods in image generation detection. For the binary classification task, we evaluate with Accuracy as the metric.
\begin{itemize}
\item \textit{\textbf{Protocol:}} Training on the combination of authentic and one edited subset, and tested on other edited subsets.
\end{itemize}

\noindent \textbf{Task 2. Edited Method Attribution.} This refers to a $n$-ways (authentic and $n-1$ editing methods) model-level attribution task. Beyond discerning between authentic and edited images, the objective is to attribute edited images to the specific editing method employed. Evaluation metrics include Accuracy, F1-score and mean Average Precision.
\begin{itemize}
\item \textit{\textbf{Protocol:}} All the authentic and edited subsets are used for training, while the test set is reserved for evaluation. 
\end{itemize}

\begin{figure*}[th]
    \centering
    \includegraphics[width=0.9\linewidth]{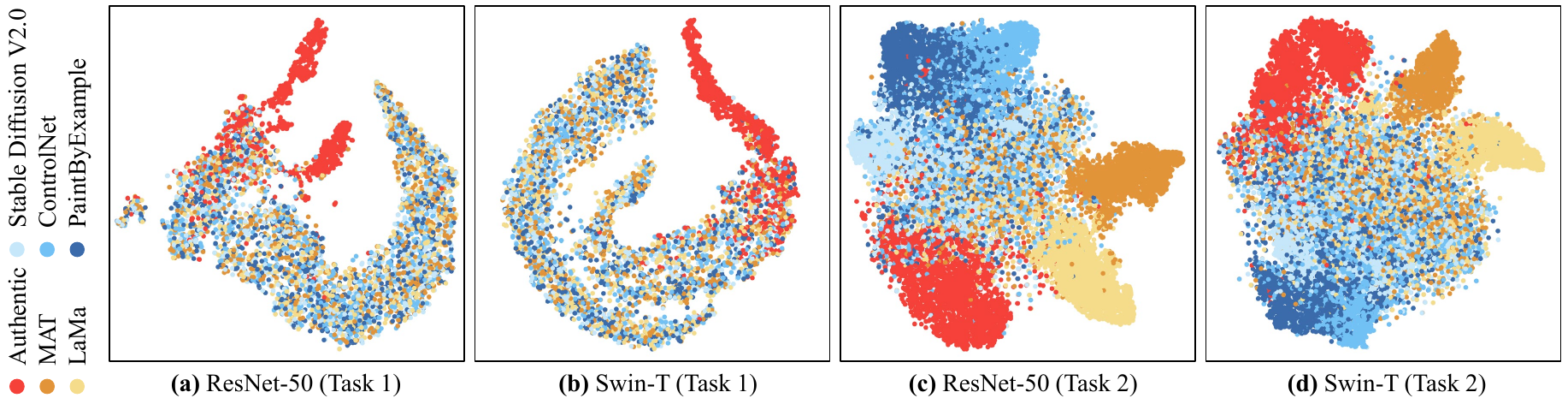}
    \caption{The t-SNE feature visualization of the authentic images and images edited by different regional editing approaches. The ResNet-50 and Swin-T are trained with protocol in Task 1 and Task 2 respectively.}
    \label{fig:tsne}
    \vspace{-1em}
\end{figure*}

\noindent \textbf{Task 3. Edited Region Localization.} This concerns a $2$-ways pixel-level segmentation task designed to distinguish between authentic and edited regions in images. For a comprehensive analysis, we introduce two protocols to analysis different editors and evaluate the existing manipulation localization methods, utilizing Intersection over Union (IoU) and F1-score as assessment metrics.

\begin{itemize}
\item \textit{\textbf{Protocol 1:}} Training on one arbitrary edited subset and tested on the other edited subsets. The authentic subset is not used since both authentic and edited regions are present within the edited subset.
\item \textit{\textbf{Protocol 2:}} Training using a combination of the MAT subset and SD-V2.0 subset, followed by testing on other subsets. The combined training set includes one GAN-based and diffusion-based editing method respectively, providing a more comprehensive assessment for image manipulation localization methods. 
\end{itemize}

\begin{table}[t]
    \centering
    \resizebox{\linewidth}{!}{
        \setlength{\tabcolsep}{1.9mm}{
            \begin{tabular}{lcccccc}
                \toprule[1.3pt]
                \multirow{2}[1]{1cm}{\textbf{Training Subset}} & \multicolumn{6}{c}{\textbf{Testing Subset} (Accuracy)}\\
                \cmidrule(r){2-7}
                & Authentic & MAT & LaMa & SD-V2.0 & ControlNet & PaintEx\\
                \midrule
                MAT & 92.2 & 88.5 & 89.1 & 85.9 & 86.3 & 85.8 \\
                LaMa & 91.9 & 89.9 & 90.0 & 87.7 & 88.1 & 87.4 \\
                StableDiff & 90.7 & 91.1 & 91.3 & 88.5 & 88.1 & 88.3 \\
                ControlNet & 86.9 & 93.6 & 94.0 & 92.4 & 91.4 & 91.5 \\
                PaintEx & 92.4 & 86.1 & 85.3 & 83.4 & 83.9 & 82.6 \\
                \bottomrule[1.3pt]
            \end{tabular}
        }
    }
    \caption{\textbf{Edited Image Classification (Task 1).} The cross-editors evaluation of ResNet-50. The authentic subset and specific subset listed in \textit{Training Subset} are used for training.}
    \label{table:task1}
\end{table}

\begin{table}[t]
    \centering
    \resizebox{\linewidth}{!}{
        \setlength{\tabcolsep}{2mm}{
            \begin{tabular}{lcccccc}
                \toprule[1.3pt]
                \multirow{2}[1]{*}{\textbf{Method}} & \multicolumn{2}{c}{\textbf{Seen Subset}} & \multicolumn{3}{c}{\textbf{Unseen Subset}} \\
                \cmidrule(r){2-3}  \cmidrule(r){4-7}
                & Authentic & SD-V2.0 & MAT & LaMa & ControlNet & PaintEx \\
                \midrule
                ResNet-18 & 89.8 & 86.5 & 79.4 & 80.6 & 81.1 & 81.9 \\
                ResNet-50 & 90.7 & 88.5 & 91.1 & \textbf{91.3} & 88.1 & 88.3 \\
                DeiT-S & 91.6 & 79.3 & 72.0 & 73.8 & 71.9 & 71.5 \\
                Swin-T & \textbf{95.4} & 87.8 & 85.5 & 85.6 & 86.1 & 85.2 \\
                \midrule
                CNNSpot & 85.8 & 73.6 & 71.3 & 72.9 & 70.7 & 69.5 \\
                F3Net & 82.3 & 68.1 & 62.4 & 61.7 & 59.8 & 60.5 \\
                GramNet & 92.7 & \textbf{93.2} & 91.5 & 90.7 & 89.0 & 88.9 \\
                Universal & 91.0 & 93.1 & \textbf{91.9} & 91.2 & \textbf{91.5} & \textbf{91.4} \\
                \bottomrule[1.3pt]
            \end{tabular}
        }
    }
    \caption{\textbf{Edited Image Classification (Task 1).} The Evaluation of various classification methods. The authentic subset and SD-V2.0 subset are used for training. The metric is Accuracy.}
    \label{table:task1_2}
    \vspace{-1em}
\end{table}

\subsubsection{Edited Image Classification}

In the setting of Task 1, we chose ResNet-50 \cite{he2016deep} as the baseline model and evaluated its performance across diverse editing subsets, as shown in Table \ref{table:task1}. Notably, the baseline model exhibits commendable generalization performance when tested on unseen subsets. There is no significant difference observed among different editing methods. Therefore, we choose the authentic subset and SD-V2.0 subset as the training sets and conduct the evaluation.

For comprehensive evaluation, we provide results of several baseline models, including ResNet-18 \cite{he2016deep}, ResNet-50 \cite{he2016deep}, DeiT-S \cite{touvron2021training} and Swin-T \cite{liu2021swin}. We extend SOTA detection methods for image generation detection to the classification task of regional editing images. It is observed that the performance of GramNet \cite{liu2020global} and Universal \cite{ojha2023towards} surpasses that of CNNSpot \cite{wang2020cnn}, F3Net \cite{qian2020thinking} and baseline. However, in Figure \ref{fig:tsne} (a) and (b), we utilize t-SNE to analyze and visualize the features of two baselines, ResNet-50 and Swin-T. An evident observation from Table \ref{table:task1_2} emerges, while the features of authentic images and edited images form a distinct classification boundary, the features of images edited using different methods do not cluster well.

\begin{figure}[t]
    \centering
    \includegraphics[width=0.95\linewidth]{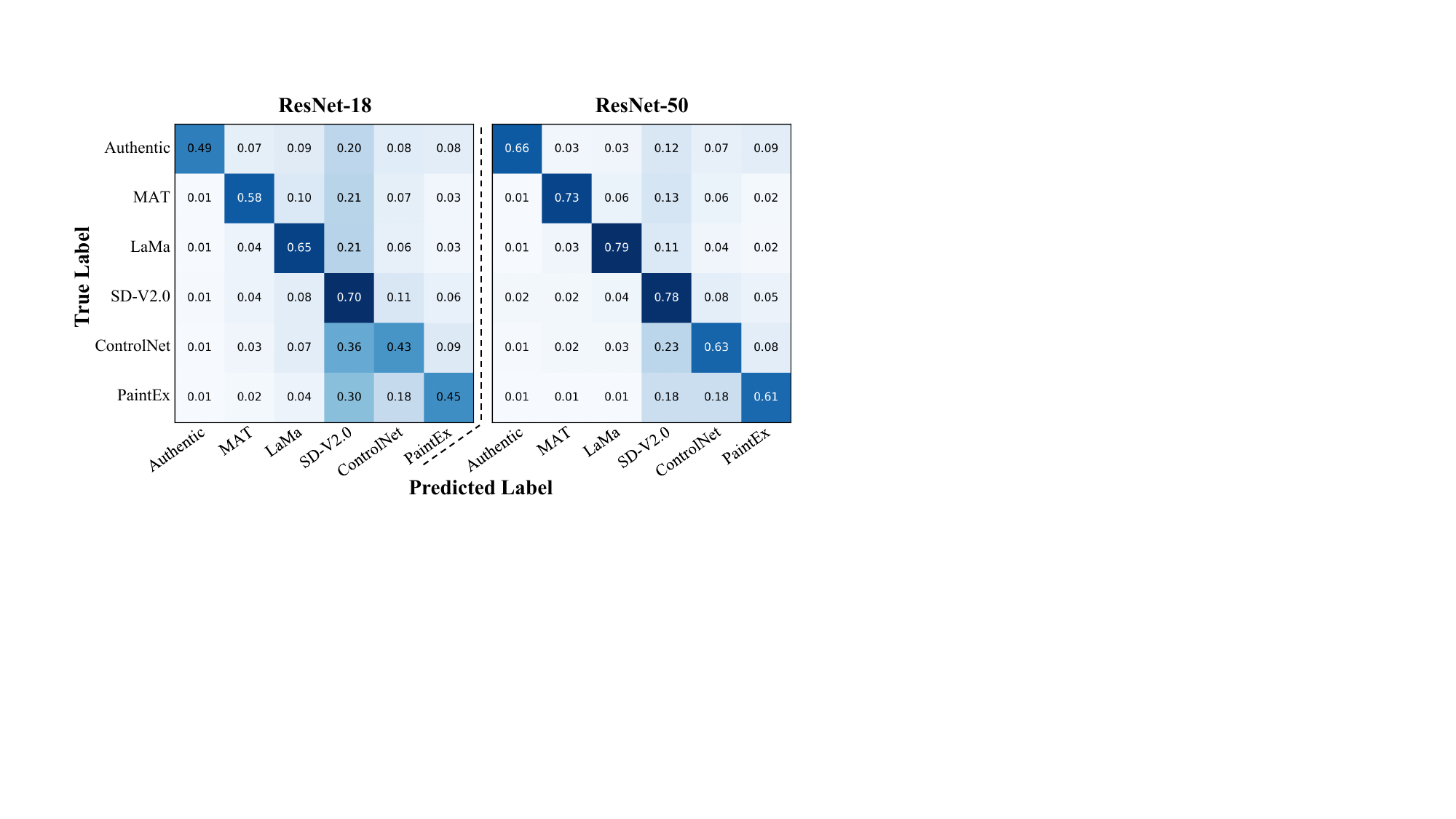}
    \caption{\textbf{Edited Method Attribution (Task 2).} The confusion matrix of ResNet-18 and ResNet-50.}
    \label{fig:confusion_matrix}
    \vspace{-1.5em}
\end{figure}

\subsubsection{Editing Method Attribution}

We expand the $2$-way classification labels of Task 1 to $n$-way attribution labels in Task 2. In addition to distinguishing between authentic and edited images, our objective is to attribute edited images to the specific editing method employed. Following the protocol in Task 2, we use the authentic subset and all edited subsets for both training and testing, constituting a closed-set attribution task. 

In addition to the classification baselines mentioned earlier, we also evaluate SOTA attribution models, including DCT-CNN \cite{frank2020leveraging}, DNA-Det \cite{yang2022deepfake}, RepMix \cite{bui2022repmix}, and POSE \cite{yang2023progressive}. The experimental results are presented in Table \ref{table:task2}. We also employ t-SNE to visualize the feature distributions of two baselines (ResNet-50 and Swin-T) under the protocol of Task 2, as shown in Figures \ref{fig:tsne} (c) and (d). Through comparison with Figures \ref{fig:tsne} (a) and (b), a crucial change is observed, where images edited by different methods cluster more distinctly. Additionally, various GAN-based methods can be well distinguished, while distinction among different diffusion-based methods is more challenging. Furthermore, in Figure \ref{fig:confusion_matrix}, we present the confusion matrices for the attribution results of ResNet-18 and ResNet-50, aligning with the observation mentioned earlier.

\begin{table}[t]
    \centering
    \resizebox{0.7\linewidth}{!}{
        \setlength{\tabcolsep}{3mm}{
            \begin{tabular}{lccc}
                \toprule[1.3pt]
                \textbf{Method} & \textbf{Accuracy} & \textbf{F1-score} & \textbf{mAP} \\
                \midrule
                ResNet-18 & 64.2 & 67.5 & 76.7 \\
                ResNet-50 & 72.6 & 73.4 & 82.8 \\
                Deit-S & 61.9 & 66.0 & 71.4 \\
                Swin-T & \textbf{74.3} & 74.7 & 82.1 \\
                \midrule
                DCT-CNN & 67.4 & 67.1 & 78.2 \\
                DNA-Det & 72.8 & 74.5 & 82.0 \\
                RepMix & 72.5 & 73.9 & \textbf{83.6} \\
                POSE & 74.1 & \textbf{75.8} & 83.1 \\
                \bottomrule[1.3pt]
            \end{tabular}
        }
    }
    \caption{\textbf{Edited Method Attribution (Task 2).} The Evaluation of various attribution methods. All the authentic and edited subsets in the training set are used for training.}
    \label{table:task2}
    \vspace{-1.5em}
\end{table}

\subsubsection{Edited Region Localization}

In the context regional edited images detection, merely distinguishing between authentic and edited images is insufficient. Locating the edited regions is a core task, and it is also the most challenging. Following the setting of \textit{Protocol 1} in Task 3, we choose Unet with EfficientNet-B4 as backbone (abbreviated as Unet-Eb4) as the baseline model and test its generalization performance across diverse editing subsets, as shown in Table \ref{table:task3}. Specifically, the baseline method demonstrates acceptable localization abilities within the seen editing subset, but its generalization capabilities significantly decrease when tested on unseen subsets. An crucial observation is that the generalization difficulty among methods with different architectures (GAN-based and diffusion-based) surpasses that between methods with the same architecture. Therefore, we design \textit{Protocol 2} for a more comprehensive and fair analysis.

To establish a comprehensive evaluation, we select classic baselines and representative image manipulation detection methods. We employ different combinations of classic segmentation models (Unet and Deeplab${}_{V3}$) and backbones (ResNet-50 and EfficientNet-B4) as baselines for the segmentation task. For Mantra-Net \cite{wu2019mantra} and SPAN \cite{hu2020span}, the core lies in their pre-trained feature extractor. Therefore, we did not retrain them on GRE training set but rather evaluated their pre-trained models on testing set, which is indicated by the asterisk. In addition, we evaluate MVSS-Net \cite{chen2021image}, PSCC-Net \cite{liu2022pscc} and SAFL-Net \cite{sun2023safl} according to \textit{Protocol 2}. The detailed experiments are presented in Table \ref{table:task3_2}.

It is worth noting that all methods exhibit acceptable localization abilities within the seen subsets. However, there is a notable lack of generalization within the unseen subsets. An important factor contributing to this phenomenon is that these methods primarily focus on non-generative forms of region editing (e.g., simple splice and copy-move). In contrast, generative regional editing approaches produce higher-quality images with less distinct boundary for edited regions. The logic and simulated characteristics of our editing pipeline further ensure that editing operations are less perceptible. This emphasizes the value of our proposed GRE dataset for the field of regional editing detection.

\begin{table}[t]
    \centering
    \resizebox{\linewidth}{!}{
        \setlength{\tabcolsep}{1.9mm}{
            \begin{tabular}{lccccc}
                \toprule[1.3pt]
                \multirow{2}[1]{1cm}{\textbf{Training Subset}} & \multicolumn{5}{c}{\textbf{Testing Subset} (IoU / F1)}\\
                \cmidrule(r){2-6}
                 & MAT & LaMa & SD-V2.0 & ControlNet & PaintEx\\
                \midrule
                MAT & \underline{76.1/85.0} & 27.7/36.9 & 2.8/4.4 & 7.1/10.6 & 4.4/6.7 \\
                LaMa & 26.0/35.9 & \underline{76.8/84.9} & 1.9/3.0 & 3.0/4.8 & 1.5/2.5 \\
                StableDiff & 15.2/21.4 & 11.2/16.2 & \underline{57.9/67.1} & 42.5/50.5 & 53.2/62.1 \\
                ControlNet & 15.2/22.3 & 5.6/8.7 & 21.8/28.2 & \underline{70.1/78.2} & 63.9/72.9 \\
                PaintEx & 13.9/19.7 & 6.0/9.0 & 33.4/41/1 & 62.1/70.2 & \underline{76.3/84.1} \\
                \bottomrule[1.3pt]
            \end{tabular}
        }
    }
    \caption{\textbf{Edited Region Localization (Task 3, Protocol 1).} The results of Unet with EfficientNet-B4 as backbone trained on different training subsets and evaluated on different testing subsets.}
    \label{table:task3}
\end{table}

\begin{table}[t]
    \centering
    \resizebox{\linewidth}{!}{
        \setlength{\tabcolsep}{1.5mm}{
            \begin{tabular}{lccccc}
                \toprule[1.3pt]
                \multirow{2}[1]{*}{\textbf{Method}} & \multicolumn{2}{c}{\textbf{Seen Subset}} & \multicolumn{3}{c}{\textbf{Unseen Subset}} \\
                \cmidrule(r){2-3}  \cmidrule(r){4-6} 
                 & MAT & SD-V2.0 & LaMa & ControlNet & PaintEx \\
                \midrule
                Unet-R50 & 72.0/80.4 & 57.9/66.1 & 29.9/38.0 & 54.7/62.9 & 62.5/70.9 \\
                Unet-Eb4 & 76.3/84.7 & 65.1/74.1 & 40.5/50.7 & 60.2/69.1 & 66.5/75.5 \\
                Deeplab${}_{V3}$-R50 & 72.6/81.4 & 61.1/70.2 & 38.6/48.2 & 59.4/68.4 & 64.8/73.9 \\
                Deeplab${}_{V3}$-Eb4 & 78.1/87.9 & 59.8/69.5 & 38.4/47.6 & 54.0/64.5 & 60.4/70.6 \\
                \midrule
                Mantra-Net* & - & - & 0.1/0.1 & 0.1/0.1 & 0.1/0.1 \\
                SPAN* & - & - & 0.1/0.1 & 0.1/0.2 & 0.1/0.1 \\
                PSCC-Net & 38.9/50.0 & 26.6/37.1 & 17.4/25.1 & 25.3/35.8 & 26.9/35.5 \\
                MVSS-Net & 63.7/73.1 & 47.6/56.8 & 25.9/33.4 & 45.2/54.0 & 52.6/62.2 \\
                SAFL-Net & 75.7/84.2 & 58.9/64.6 & 35.6/41.1 & 61.0/67.5 & 65.4/74.9 \\
                \bottomrule[1.3pt]
            \end{tabular}
        }
    }
    \caption{\textbf{Edited Region Localization (Task 3, Protocol 2).} The results of different methods trained on the combination of MAT subset and SD-V2.0 subset. IoU and F1-score are reported.}
    \label{table:task3_2}
    \vspace{-1.5em}
\end{table}

%% file: sec/5_conclusion.tex
\section{Conclusion}

We construct GRE, a novel and expansive benchmark designed for generative regional editing detection and analysis. Compared with existing datasets for regional editing detection, GRE stands out by curating a diverse collection of real-world images, designing a simulated editing pipeline, and including a variety of generative regional editing approaches. We also introduce three practical applications, including edited image classification, edited method attribution, and edited region localization, providing a comprehensive analysis of regional editing within the context of emerging scenarios. Our future work involves continually enriching GRE by incorporating new editing methods and novel large models into our pipeline. We aim for GRE to foster innovation and progress in this evolving field.

%% file: main.bbl
\begin{thebibliography}{52}
\providecommand{\natexlab}[1]{#1}
\providecommand{\url}[1]{\texttt{#1}}
\expandafter\ifx\csname urlstyle\endcsname\relax
  \providecommand{\doi}[1]{doi: #1}\else
  \providecommand{\doi}{doi: \begingroup \urlstyle{rm}\Url}\fi

\bibitem[cha()]{chatgpt}
Openai chatgpt.
\newblock \url{https://chat.openai.com}.

\bibitem[fli()]{flickr}
Flickr.
\newblock \url{https://www.flickr.com}.

\bibitem[pho()]{photoshop}
Adobe photoshop.
\newblock \url{https://www.adobe.com/products/photoshop.html}.

\bibitem[Agustsson and Timofte(2017)]{agustsson2017ntire}
Eirikur Agustsson and Radu Timofte.
\newblock Ntire 2017 challenge on single image super-resolution: Dataset and study.
\newblock In \emph{Proceedings of the IEEE conference on computer vision and pattern recognition workshops}, pages 126--135, 2017.

\bibitem[Bird and Lotfi(2023)]{bird2023cifake}
Jordan~J Bird and Ahmad Lotfi.
\newblock Cifake: Image classification and explainable identification of ai-generated synthetic images.
\newblock \emph{arXiv preprint arXiv:2303.14126}, 2023.

\bibitem[Bui et~al.(2022)Bui, Yu, and Collomosse]{bui2022repmix}
Tu Bui, Ning Yu, and John Collomosse.
\newblock Repmix: Representation mixing for robust attribution of synthesized images.
\newblock In \emph{European Conference on Computer Vision}, pages 146--163. Springer, 2022.

\bibitem[Chen et~al.(2021)Chen, Dong, Ji, Cao, and Li]{chen2021image}
Xinru Chen, Chengbo Dong, Jiaqi Ji, Juan Cao, and Xirong Li.
\newblock Image manipulation detection by multi-view multi-scale supervision.
\newblock In \emph{Proceedings of the IEEE/CVF International Conference on Computer Vision}, pages 14185--14193, 2021.

\bibitem[Dong et~al.(2013)Dong, Wang, and Tan]{dong2013casia}
Jing Dong, Wei Wang, and Tieniu Tan.
\newblock Casia image tampering detection evaluation database.
\newblock In \emph{2013 IEEE China Summit and International Conference on Signal and Information Processing}, pages 422--426. IEEE, 2013.

\bibitem[Frank et~al.(2020)Frank, Eisenhofer, Sch{\"o}nherr, Fischer, Kolossa, and Holz]{frank2020leveraging}
Joel Frank, Thorsten Eisenhofer, Lea Sch{\"o}nherr, Asja Fischer, Dorothea Kolossa, and Thorsten Holz.
\newblock Leveraging frequency analysis for deep fake image recognition.
\newblock In \emph{International conference on machine learning}, pages 3247--3258. PMLR, 2020.

\bibitem[Guan et~al.(2019)Guan, Kozak, Robertson, Lee, Yates, Delgado, Zhou, Kheyrkhah, Smith, and Fiscus]{guan2019mfc}
Haiying Guan, Mark Kozak, Eric Robertson, Yooyoung Lee, Amy~N Yates, Andrew Delgado, Daniel Zhou, Timothee Kheyrkhah, Jeff Smith, and Jonathan Fiscus.
\newblock Mfc datasets: Large-scale benchmark datasets for media forensic challenge evaluation.
\newblock In \emph{2019 IEEE Winter Applications of Computer Vision Workshops (WACVW)}, pages 63--72. IEEE, 2019.

\bibitem[He et~al.(2016)He, Zhang, Ren, and Sun]{he2016deep}
Kaiming He, Xiangyu Zhang, Shaoqing Ren, and Jian Sun.
\newblock Deep residual learning for image recognition.
\newblock In \emph{Proceedings of the IEEE conference on computer vision and pattern recognition}, pages 770--778, 2016.

\bibitem[Ho et~al.(2020)Ho, Jain, and Abbeel]{DDPM}
Jonathan Ho, Ajay Jain, and Pieter Abbeel.
\newblock Denoising diffusion probabilistic models.
\newblock pages 6840--6851, 2020.

\bibitem[Hu et~al.(2020)Hu, Zhang, Jiang, Chaudhuri, Yang, and Nevatia]{hu2020span}
Xuefeng Hu, Zhihan Zhang, Zhenye Jiang, Syomantak Chaudhuri, Zhenheng Yang, and Ram Nevatia.
\newblock Span: Spatial pyramid attention network for image manipulation localization.
\newblock In \emph{European conference on computer vision}, pages 312--328. Springer, 2020.

\bibitem[Kirillov et~al.(2023)Kirillov, Mintun, Ravi, Mao, Rolland, Gustafson, Xiao, Whitehead, Berg, Lo, et~al.]{kirillov2023segment}
Alexander Kirillov, Eric Mintun, Nikhila Ravi, Hanzi Mao, Chloe Rolland, Laura Gustafson, Tete Xiao, Spencer Whitehead, Alexander~C Berg, Wan-Yen Lo, et~al.
\newblock Segment anything.
\newblock \emph{arXiv preprint arXiv:2304.02643}, 2023.

\bibitem[Li et~al.(2023)Li, Li, Savarese, and Hoi]{li2023blip}
Junnan Li, Dongxu Li, Silvio Savarese, and Steven Hoi.
\newblock Blip-2: Bootstrapping language-image pre-training with frozen image encoders and large language models.
\newblock \emph{arXiv preprint arXiv:2301.12597}, 2023.

\bibitem[Li et~al.(2022)Li, Lin, Zhou, Qi, Wang, and Jia]{li2022mat}
Wenbo Li, Zhe Lin, Kun Zhou, Lu Qi, Yi Wang, and Jiaya Jia.
\newblock Mat: Mask-aware transformer for large hole image inpainting.
\newblock In \emph{Proceedings of the IEEE/CVF conference on computer vision and pattern recognition}, pages 10758--10768, 2022.

\bibitem[Lin et~al.(2014)Lin, Maire, Belongie, Hays, Perona, Ramanan, Doll{\'a}r, and Zitnick]{lin2014microsoft}
Tsung-Yi Lin, Michael Maire, Serge Belongie, James Hays, Pietro Perona, Deva Ramanan, Piotr Doll{\'a}r, and C~Lawrence Zitnick.
\newblock Microsoft coco: Common objects in context.
\newblock In \emph{European conference on computer vision}, pages 740--755. Springer, 2014.

\bibitem[Liu et~al.(2020{\natexlab{a}})Liu, Wang, Wang, and Ordonez]{liu2020visual}
Fuxiao Liu, Yinghan Wang, Tianlu Wang, and Vicente Ordonez.
\newblock Visual news: Benchmark and challenges in news image captioning.
\newblock \emph{arXiv preprint arXiv:2010.03743}, 2020{\natexlab{a}}.

\bibitem[Liu et~al.(2022)Liu, Liu, Chen, and Liu]{liu2022pscc}
Xiaohong Liu, Yaojie Liu, Jun Chen, and Xiaoming Liu.
\newblock Pscc-net: Progressive spatio-channel correlation network for image manipulation detection and localization.
\newblock \emph{IEEE Transactions on Circuits and Systems for Video Technology}, 32\penalty0 (11):\penalty0 7505--7517, 2022.

\bibitem[Liu et~al.(2020{\natexlab{b}})Liu, Qi, and Torr]{liu2020global}
Zhengzhe Liu, Xiaojuan Qi, and Philip~HS Torr.
\newblock Global texture enhancement for fake face detection in the wild.
\newblock In \emph{Proceedings of the IEEE/CVF conference on computer vision and pattern recognition}, pages 8060--8069, 2020{\natexlab{b}}.

\bibitem[Liu et~al.(2021)Liu, Lin, Cao, Hu, Wei, Zhang, Lin, and Guo]{liu2021swin}
Ze Liu, Yutong Lin, Yue Cao, Han Hu, Yixuan Wei, Zheng Zhang, Stephen Lin, and Baining Guo.
\newblock Swin transformer: Hierarchical vision transformer using shifted windows.
\newblock In \emph{Proceedings of the IEEE/CVF international conference on computer vision}, pages 10012--10022, 2021.

\bibitem[Lu et~al.(2023)Lu, Huang, Bai, Liu, Qu, and Ouyang]{lu2023seeing}
Zeyu Lu, Di Huang, Lei Bai, Xihui Liu, Jingjing Qu, and Wanli Ouyang.
\newblock Seeing is not always believing: A quantitative study on human perception of ai-generated images.
\newblock \emph{arXiv preprint arXiv:2304.13023}, 2023.

\bibitem[Mahfoudi et~al.(2019)Mahfoudi, Tajini, Retraint, Morain-Nicolier, Dugelay, and Marc]{mahfoudi2019defacto}
Ga{\"e}l Mahfoudi, Badr Tajini, Florent Retraint, Frederic Morain-Nicolier, Jean~Luc Dugelay, and PIC Marc.
\newblock Defacto: image and face manipulation dataset.
\newblock In \emph{2019 27Th european signal processing conference (EUSIPCO)}, pages 1--5. IEEE, 2019.

\bibitem[Novozamsky et~al.(2020)Novozamsky, Mahdian, and Saic]{novozamsky2020imd2020}
Adam Novozamsky, Babak Mahdian, and Stanislav Saic.
\newblock Imd2020: a large-scale annotated dataset tailored for detecting manipulated images.
\newblock In \emph{Proceedings of the IEEE/CVF Winter Conference on Applications of Computer Vision Workshops}, pages 71--80, 2020.

\bibitem[Ojha et~al.(2023)Ojha, Li, and Lee]{ojha2023towards}
Utkarsh Ojha, Yuheng Li, and Yong~Jae Lee.
\newblock Towards universal fake image detectors that generalize across generative models.
\newblock In \emph{Proceedings of the IEEE/CVF Conference on Computer Vision and Pattern Recognition}, pages 24480--24489, 2023.

\bibitem[Ouyang et~al.(2022)Ouyang, Wu, Jiang, Almeida, Wainwright, Mishkin, Zhang, Agarwal, Slama, Ray, et~al.]{ouyang2022training}
Long Ouyang, Jeffrey Wu, Xu Jiang, Diogo Almeida, Carroll Wainwright, Pamela Mishkin, Chong Zhang, Sandhini Agarwal, Katarina Slama, Alex Ray, et~al.
\newblock Training language models to follow instructions with human feedback.
\newblock \emph{Advances in Neural Information Processing Systems}, 35:\penalty0 27730--27744, 2022.

\bibitem[Podell et~al.(2023)Podell, English, Lacey, Blattmann, Dockhorn, M{\"u}ller, Penna, and Rombach]{podell2023sdxl}
Dustin Podell, Zion English, Kyle Lacey, Andreas Blattmann, Tim Dockhorn, Jonas M{\"u}ller, Joe Penna, and Robin Rombach.
\newblock Sdxl: Improving latent diffusion models for high-resolution image synthesis.
\newblock \emph{arXiv preprint arXiv:2307.01952}, 2023.

\bibitem[Qian et~al.(2020)Qian, Yin, Sheng, Chen, and Shao]{qian2020thinking}
Yuyang Qian, Guojun Yin, Lu Sheng, Zixuan Chen, and Jing Shao.
\newblock Thinking in frequency: Face forgery detection by mining frequency-aware clues.
\newblock In \emph{European conference on computer vision}, pages 86--103. Springer, 2020.

\bibitem[Radford et~al.(2021)Radford, Kim, Hallacy, Ramesh, Goh, Agarwal, Sastry, Askell, Mishkin, Clark, et~al.]{radford2021learning}
Alec Radford, Jong~Wook Kim, Chris Hallacy, Aditya Ramesh, Gabriel Goh, Sandhini Agarwal, Girish Sastry, Amanda Askell, Pamela Mishkin, Jack Clark, et~al.
\newblock Learning transferable visual models from natural language supervision.
\newblock In \emph{International conference on machine learning}, pages 8748--8763. PMLR, 2021.

\bibitem[Rombach et~al.(2022)Rombach, Blattmann, Lorenz, Esser, and Ommer]{rombach2022high}
Robin Rombach, Andreas Blattmann, Dominik Lorenz, Patrick Esser, and Bj{\"o}rn Ommer.
\newblock High-resolution image synthesis with latent diffusion models.
\newblock In \emph{Proceedings of the IEEE/CVF conference on computer vision and pattern recognition}, pages 10684--10695, 2022.

\bibitem[Saharia et~al.(2022)Saharia, Chan, Saxena, Li, Whang, Denton, Ghasemipour, Gontijo~Lopes, Karagol~Ayan, Salimans, et~al.]{saharia2022photorealistic}
Chitwan Saharia, William Chan, Saurabh Saxena, Lala Li, Jay Whang, Emily~L Denton, Kamyar Ghasemipour, Raphael Gontijo~Lopes, Burcu Karagol~Ayan, Tim Salimans, et~al.
\newblock Photorealistic text-to-image diffusion models with deep language understanding.
\newblock \emph{Advances in Neural Information Processing Systems}, 35:\penalty0 36479--36494, 2022.

\bibitem[Shi and Malik(2000)]{shi2000normalized}
Jianbo Shi and Jitendra Malik.
\newblock Normalized cuts and image segmentation.
\newblock \emph{IEEE Transactions on pattern analysis and machine intelligence}, 22\penalty0 (8):\penalty0 888--905, 2000.

\bibitem[Song et~al.(2020)Song, Sohl-Dickstein, Kingma, Kumar, Ermon, and Poole]{Score-based}
Yang Song, Jascha Sohl-Dickstein, Diederik~P Kingma, Abhishek Kumar, Stefano Ermon, and Ben Poole.
\newblock Score-based generative modeling through stochastic differential equations.
\newblock 2020.

\bibitem[Sun et~al.(2023)Sun, Jiang, Wang, Li, and Cao]{sun2023safl}
Zhihao Sun, Haoran Jiang, Danding Wang, Xirong Li, and Juan Cao.
\newblock Safl-net: Semantic-agnostic feature learning network with auxiliary plugins for image manipulation detection.
\newblock In \emph{Proceedings of the IEEE/CVF International Conference on Computer Vision}, pages 22424--22433, 2023.

\bibitem[Suvorov et~al.(2022)Suvorov, Logacheva, Mashikhin, Remizova, Ashukha, Silvestrov, Kong, Goka, Park, and Lempitsky]{suvorov2022resolution}
Roman Suvorov, Elizaveta Logacheva, Anton Mashikhin, Anastasia Remizova, Arsenii Ashukha, Aleksei Silvestrov, Naejin Kong, Harshith Goka, Kiwoong Park, and Victor Lempitsky.
\newblock Resolution-robust large mask inpainting with fourier convolutions.
\newblock In \emph{Proceedings of the IEEE/CVF winter conference on applications of computer vision}, pages 2149--2159, 2022.

\bibitem[Timofte et~al.(2017)Timofte, Agustsson, Van~Gool, Yang, and Zhang]{timofte2017ntire}
Radu Timofte, Eirikur Agustsson, Luc Van~Gool, Ming-Hsuan Yang, and Lei Zhang.
\newblock Ntire 2017 challenge on single image super-resolution: Methods and results.
\newblock In \emph{Proceedings of the IEEE conference on computer vision and pattern recognition workshops}, pages 114--125, 2017.

\bibitem[Touvron et~al.(2021)Touvron, Cord, Douze, Massa, Sablayrolles, and J{\'e}gou]{touvron2021training}
Hugo Touvron, Matthieu Cord, Matthijs Douze, Francisco Massa, Alexandre Sablayrolles, and Herv{\'e} J{\'e}gou.
\newblock Training data-efficient image transformers \& distillation through attention.
\newblock In \emph{International conference on machine learning}, pages 10347--10357. PMLR, 2021.

\bibitem[Verdoliva et~al.()Verdoliva, Cozzolino, and Nagano]{verdoliva2022}
Luisa Verdoliva, Davide Cozzolino, and Koki Nagano.
\newblock 2022 ieee image and video processing cup synthetic image detection.

\bibitem[Wang et~al.(2020)Wang, Wang, Zhang, Owens, and Efros]{wang2020cnn}
Sheng-Yu Wang, Oliver Wang, Richard Zhang, Andrew Owens, and Alexei~A Efros.
\newblock Cnn-generated images are surprisingly easy to spot... for now.
\newblock In \emph{Proceedings of the IEEE/CVF conference on computer vision and pattern recognition}, pages 8695--8704, 2020.

\bibitem[Wang et~al.(2023)Wang, Huang, and Hong]{wang2023benchmarking}
Yabin Wang, Zhiwu Huang, and Xiaopeng Hong.
\newblock Benchmarking deepart detection.
\newblock \emph{arXiv preprint arXiv:2302.14475}, 2023.

\bibitem[Wen et~al.(2016)Wen, Zhu, Subramanian, Ng, Shen, and Winkler]{wen2016coverage}
Bihan Wen, Ye Zhu, Ramanathan Subramanian, Tian-Tsong Ng, Xuanjing Shen, and Stefan Winkler.
\newblock Coverage—a novel database for copy-move forgery detection.
\newblock In \emph{2016 IEEE international conference on image processing (ICIP)}, pages 161--165. IEEE, 2016.

\bibitem[Wu et~al.(2019)Wu, AbdAlmageed, and Natarajan]{wu2019mantra}
Yue Wu, Wael AbdAlmageed, and Premkumar Natarajan.
\newblock Mantra-net: Manipulation tracing network for detection and localization of image forgeries with anomalous features.
\newblock In \emph{Proceedings of the IEEE/CVF Conference on Computer Vision and Pattern Recognition}, pages 9543--9552, 2019.

\bibitem[Xu et~al.(2023)Xu, Wang, Meng, Mi, Yuan, and Yan]{xu2023exposing}
Qiang Xu, Hao Wang, Laijin Meng, Zhongjie Mi, Jianye Yuan, and Hong Yan.
\newblock Exposing fake images generated by text-to-image diffusion models.
\newblock \emph{Pattern Recognition Letters}, 2023.

\bibitem[Yang et~al.(2023{\natexlab{a}})Yang, Gu, Zhang, Zhang, Chen, Sun, Chen, and Wen]{yang2023paint}
Binxin Yang, Shuyang Gu, Bo Zhang, Ting Zhang, Xuejin Chen, Xiaoyan Sun, Dong Chen, and Fang Wen.
\newblock Paint by example: Exemplar-based image editing with diffusion models.
\newblock In \emph{Proceedings of the IEEE/CVF Conference on Computer Vision and Pattern Recognition}, pages 18381--18391, 2023{\natexlab{a}}.

\bibitem[Yang et~al.(2022)Yang, Huang, Cao, Li, and Li]{yang2022deepfake}
Tianyun Yang, Ziyao Huang, Juan Cao, Lei Li, and Xirong Li.
\newblock Deepfake network architecture attribution.
\newblock In \emph{Proceedings of the AAAI Conference on Artificial Intelligence}, pages 4662--4670, 2022.

\bibitem[Yang et~al.(2023{\natexlab{b}})Yang, Cao, Wang, and Xu]{yang2023fingerprints}
Tianyun Yang, Juan Cao, Danding Wang, and Chang Xu.
\newblock Fingerprints of generative models in the frequency domain.
\newblock \emph{arXiv preprint arXiv:2307.15977}, 2023{\natexlab{b}}.

\bibitem[Yang et~al.(2023{\natexlab{c}})Yang, Wang, Tang, Zhao, Cao, and Tang]{yang2023progressive}
Tianyun Yang, Danding Wang, Fan Tang, Xinying Zhao, Juan Cao, and Sheng Tang.
\newblock Progressive open space expansion for open-set model attribution.
\newblock In \emph{Proceedings of the IEEE/CVF Conference on Computer Vision and Pattern Recognition}, pages 15856--15865, 2023{\natexlab{c}}.

\bibitem[Yu et~al.(2022)Yu, Xu, Koh, Luong, Baid, Wang, Vasudevan, Ku, Yang, Ayan, et~al.]{yu2022scaling}
Jiahui Yu, Yuanzhong Xu, Jing~Yu Koh, Thang Luong, Gunjan Baid, Zirui Wang, Vijay Vasudevan, Alexander Ku, Yinfei Yang, Burcu~Karagol Ayan, et~al.
\newblock Scaling autoregressive models for content-rich text-to-image generation.
\newblock \emph{arXiv preprint arXiv:2206.10789}, 2\penalty0 (3):\penalty0 5, 2022.

\bibitem[Zhang et~al.(2021)Zhang, Yin, Fang, Li, Duan, Wu, Sun, Tian, Wu, and Wang]{zhang2021ernie}
Han Zhang, Weichong Yin, Yewei Fang, Lanxin Li, Boqiang Duan, Zhihua Wu, Yu Sun, Hao Tian, Hua Wu, and Haifeng Wang.
\newblock Ernie-vilg: Unified generative pre-training for bidirectional vision-language generation.
\newblock \emph{arXiv preprint arXiv:2112.15283}, 2021.

\bibitem[Zhang and Agrawala(2023)]{zhang2023adding}
Lvmin Zhang and Maneesh Agrawala.
\newblock Adding conditional control to text-to-image diffusion models.
\newblock \emph{arXiv preprint arXiv:2302.05543}, 2023.

\bibitem[Zhang et~al.(2019)Zhang, Chen, Ng, and Koltun]{zhang2019zoom}
Xuaner Zhang, Qifeng Chen, Ren Ng, and Vladlen Koltun.
\newblock Zoom to learn, learn to zoom.
\newblock In \emph{Proceedings of the IEEE/CVF Conference on Computer Vision and Pattern Recognition}, pages 3762--3770, 2019.

\bibitem[Zhu et~al.(2023)Zhu, Chen, Yan, Huang, Lin, Li, Tu, Hu, Hu, and Wang]{zhu2023genimage}
Mingjian Zhu, Hanting Chen, Qiangyu Yan, Xudong Huang, Guanyu Lin, Wei Li, Zhijun Tu, Hailin Hu, Jie Hu, and Yunhe Wang.
\newblock Genimage: A million-scale benchmark for detecting ai-generated image.
\newblock \emph{arXiv preprint arXiv:2306.08571}, 2023.

\end{thebibliography}
